%% file: manuscript.tex
\documentclass{article} 
\usepackage{iclr2020_conference,times}
\input{math_commands.tex}

\usepackage{hyperref}
\usepackage{url}

\usepackage{lipsum}
\usepackage{amsmath}
\usepackage{xcolor}
\usepackage{graphicx}
\usepackage[ruled]{algorithm2e}

\newcommand{\bias}{\Phi}

\usepackage{hyperref}
\hypersetup{colorlinks,linkcolor={red},citecolor={red},urlcolor={red}}  

\title{Reinforcement Learning through \\ Active Inference}

\author{Alexander Tschantz \\
Sackler Centre for Consciousness Science\\
Evolutionary \& Adaptive Systems Research Group\\
University of Sussex \\
Brighton, UK\\
\texttt{tschantz.alec@gmail.com} \\
\And
Beren Millidge \\
University of Edinburgh \\
Edinburgh, UK\\
\texttt{s1686853@sms.ed.ac.uk} \\
\And
Anil K. Seth \\
Sackler Centre for Consciousness Science\\
Evolutionary \& Adaptive Systems Research Group\\
University of Sussex \\
Brighton, UK\\
Canadian Institute for Advanced Research\\
\And
Christopher L. Buckley \\
Evolutionary \& Adaptive Systems Research Group\\
University of Sussex \\
Brighton, UK\\
}

\iclrfinalcopy
\begin{document}
\maketitle


\begin{abstract}
   The central tenet of reinforcement learning (RL) is that agents seek to maximize the sum of cumulative rewards. 
   In contrast, active inference, an emerging framework within cognitive and computational neuroscience, proposes that agents act to maximize the evidence for a biased generative model. 
   Here, we illustrate how ideas from active inference can augment traditional RL approaches by (i) furnishing an inherent balance of exploration and exploitation, and (ii) providing a more flexible conceptualization of reward. 
   Inspired by active inference, we develop and implement a novel objective for decision making, which we term the \textit{free energy of the expected future}. 
   We demonstrate that the resulting algorithm successfully balances exploration and exploitation, simultaneously achieving robust performance on several challenging RL benchmarks with sparse, well-shaped, and no rewards.
\end{abstract}


\section{Introduction}
\label{Introduction}

Both biological and artificial agents must learn to make adaptive decisions in unknown environments. 
In the field of reinforcement learning (RL), agents aim to learn a policy that maximises the sum of expected rewards \citep{sutton1998introduction}. 
This approach has demonstrated impressive results in domains such as simulated games \citep{mnih2015human,silver2017mastering}, robotics \citep{polydoros2017survey,nagabandi2019deep} and industrial applications \citep{meyes2017motion}.
 
In contrast, active inference \citep{friston2016active,friston_active_2015,friston2012active,friston2009reinforcement} - an emerging framework from cognitive and computational neuroscience - suggests that agents select actions in order to maximise the evidence for a model that is biased towards an agent's preferences.
This framework extends influential theories of Bayesian perception and learning \citep{knill2004bayesian, l2008bayesian} to incorporate probabilistic decision making, and comes equipped with a biologically plausible process theory \citep{friston2017active} that enjoys considerable empirical support \citep{friston_predictive_2009}.

Although active inference and RL have their roots in different disciplines, both frameworks have converged upon similar solutions to the problem of learning adaptive behaviour. 
For instance, both frameworks highlight the importance of learning probabilistic models, performing inference and efficient planning. 
This leads to a natural question: can insights from active inference inform the development of novel RL algorithms?

Conceptually, there are several ways in which active inference can inform and potentially enhance the field of RL. 
First, active inference suggests that agents embody a generative model of their preferred environment and seek to maximise the evidence for this model. 
In this context, rewards are cast as prior probabilities over observations, and success is measured in terms of the divergence between preferred and expected outcomes. 
Formulating preferences as prior probabilities enables greater flexibility when specifying an agent's goals \citep{friston2012active,friston2019particularphysics}, provides a principled (i.e. Bayesian) method for learning preferences \citep{sajid2019demystifying}, and is consistent with recent neurophysiological data demonstrating the distributional nature of reward representations \citep{dabney2020distributional}. 
Second, reformulating reward maximisation as maximizing model evidence naturally encompasses both exploration and exploitation under a single objective, obviating the need for adding ad-hoc exploratory terms to existing objectives.
Moreover, as we will show, active inference subsumes a number of established RL formalisms, indicating a potentially unified framework for adaptive decision-making under uncertainty.

Translating these conceptual insights into practical benefits for RL has proven challenging. 
Current implementations of active inference have generally been confined to discrete state spaces and toy problems \citep{friston_active_2015,friston_active_2017,friston_active_2017-1} (although see \citep{tschantz_scaling_2019,millidge_deep_2019, catal_bayesian_2019}). 
Therefore, it has not yet been possible to evaluate the effectiveness of active inference in challenging environments; as a result, active inference has not yet been widely taken up within the RL community. 

In this paper, we consider active inference in the context of decision making\footnote{A full treatment of active inference would consider inference and learning, see \citep{buckley2017free} for an overview.}. 
We propose and implement a novel objective function for active inference - the \textit{free energy of the expected future} - and show that this quantity provides a tractable bound on established RL objectives. 
We evaluate the performance of this algorithm on a selection of challenging continuous control tasks. 
We show strong performance on environments with sparse, well-shaped, and no rewards, demonstrating our algorithm's ability to effectively balance exploration and exploitation. 
Altogether, our results indicate that active inference provides a promising complement to current RL methods.


\section{Active Inference}
\label{active_Inference}

Both active inference and RL can be formulated in the context of a partially observed Markov decision process POMDPs \citep{murphy1982survey}.
At each time step $t$, the true state of the environment $\mathbf{s}_t$ evolves according to the stochastic transition dynamics $\mathbf{\rs}_t \sim \mathrm{p}(\mathbf{s}_t|\mathbf{s}_{t-1}, \mathbf{a}_{t-1})$, where $\mathbf{a} \in \mathbb{R}^{d_{a}}$ denotes an agent's actions. 
Agents do not necessarily have access to the true state of the environment, but may instead receive observations $\ro_t \in \mathbb{R}^{d_{o}}$, which are generated according to $\ro_t \sim \mathrm{p}(\ro_t|\mathbf{s}_t)$. 
In this case, agents must operate on beliefs $\rs_t \in \mathbb{R}^{d_{s}}$ about the true state of the environment $\mathbf{s}_t$. 
Finally, the environment generates rewards $\rr_t$ according to $\rr_t \sim \mathrm{p}(\rr_t|\mathbf{s}_t)$\footnote{We use $\mathbf{x}$ and $\mathrm{p}(\cdot)$ to denote the generative process and $\rx$ and $p(\cdot)$ to denote the agent's generative model.}. 

The goal of RL is to learn a policy that maximises the expected sum of rewards $\E[\sum_{t=0}^{\infty} \gamma^{t} \rr_{t}]$ \citep{sutton1998introduction}. 
In contrast, the goal of active inference is to maximise the Bayesian model evidence for an agent's generative model $p^{\bias}(\ro, \rs, \theta)$, where $\theta \in \Theta$ denote model parameters.

Crucially, active inference allows that an agent's generative model can be biased towards favourable states of affairs \citep{friston_free_2019}.
In other words, the model assigns probability to the parts of observation space that are both likely \textit{and} beneficial for an agent's success.  
We use the notation $p^{\bias}(\cdot)$ to represent an arbitrary distribution encoding the agent's preferences.

Given a generative model, agents can perform approximate Bayesian inference by encoding an arbitrary distribution $q(\rs, \theta)$ and minimising \textit{variational free energy} $\mathcal{F} = \KL \Big( q(\rs, \theta) \Vert p^{\bias}(\ro, \rs, \theta) \Big)$. 
When observations $\ro$ are known, $\mathcal{F}$ can be minimized through standard variational methods \citep{bishop2006pattern,buckley2017free}, causing $q(\rs, \theta)$ to tend towards the true posterior $p(\rs, \theta|\ro)$. 
Note that treating model parameters $\theta$ as random variables casts learning as a process of inference \citep{blundell_weight_2015}.

In the current context, agents additionally maintain beliefs over policies $\pi = \{a_0,...,a_T \}$, which are themselves random variables. 
Policy selection is then implemented by identifying $q(\pi)$ that minimizes $\mathcal{F}$, thus casting policy selection as a process of approximate inference \citep{friston_active_2015}.
While the standard free energy functional $\mathcal{F}$ is generally defined for a single time point $t$, $\pi$ refers to a temporal sequence of variables. 
Therefore, we augment the free energy functional $\mathcal{F}$ to encompass future variables, leading to the \textit{free energy of the expected future} $\mathcal{\tilde{F}}$.
This quantity measures the KL-divergence between a sequence of beliefs about future variables and an agent's biased generative model.  

The goal is now to infer $q(\pi)$ in order to minimise $\mathcal{\tilde{F}}$. 
We demonstrate that the resulting scheme naturally encompasses both exploration and exploitation, thereby suggesting a deep relationship between inference, learning and decision making.


\section{Free energy of the expected future}
\label{expected_future}

Let $\rx_{t:T}$ denote a sequence of variables through time, $\rx_{t:T} = \{\rx_t , ... , \rx_T \}$. 
We wish to minimize the free energy of the expected future $\mathcal{\tilde{F}}$, which is defined as:
\begin{equation}
\label{eq:feef}
   \mathcal{\tilde{F}} = \KL \Big( q(\ro_{0:T}, \rs_{0:T}, \theta, \pi) \Vert p^{\bias}(\ro_{0:T}, \rs_{0:T}, \theta) \Big)
\end{equation}

where $q(\ro_{t:T}, \rs_{t:T}, \theta, \pi)$ represents an agent's beliefs about future variables, and $p^{\bias}(\ro_{t:T}, \rs_{t:T}, \theta)$ represents an agent's biased generative model. 
Note that the beliefs about future variables include beliefs about future observations, $\ro_{t:T}$, which are unknown and thus treated as random variables\footnote{For readers familiar with the active inference framework, we highlight that the \textit{free energy of the expected future} differs from \textit{expected free energy} \citep{friston_active_2015}. We leave a discussion of the relative merits to future work.}.

In order to find $q(\pi)$ which minimizes $\mathcal{\tilde{F}}$ we note that (see Appendix \ref{ap:opt-policy}):
\begin{equation}
\begin{aligned}
\label{eq:control_eq}
         \mathcal{\tilde{F}} = 0 \Rightarrow \KL \Big( q(\pi) \ \Vert \big(- e^{-\mathcal{\tilde{F_{\pi}}}} \big) \Big) = 0 \\
\end{aligned}
\end{equation}
where
\begin{equation}
\begin{aligned}
\label{eq:feef_pi}
\mathcal{\tilde{F_{\pi}}} = \KL \Big( q(\ro_{0:T}, \rs_{0:T}, \theta | \pi)  \ \Vert \ p^{\bias}(\ro_{0:T}, \rs_{0:T}, \theta) \Big) 
\end{aligned}
\end{equation}

Thus, the free energy of the expected future is minimized when $q(\pi) = \sigma(-\mathcal{\tilde{F_{\pi}}})$, or in other words, policies are more likely when they minimise $\mathcal{\tilde{F_{\pi}}}$.


\subsection{Exploration \& exploitation}
\label{expl_exploit}

In order to provide an intuition for what minimizing $\mathcal{\tilde{F_{\pi}}}$ entails, we factorize the agent's generative models as $p^{\bias}(\ro_{0:T}, \rs_{0:T}, \theta) = p(\rs_{0:T}, \theta| \ro_{0:T}) p^{\bias}(\ro_{0:T})$, implying that the model is only biased in its beliefs over observations. 
To retain consistency with RL nomenclature, we treat `rewards' $\rr$ as a separate observation modality, such that $p^{\bias}(\ro_{t:T})$ specifies a distribution over preferred rewards. 
We describe our implementation of $p^{\bias}(\ro_{t:T})$ in Appendix \ref{ap:model-details}.
In a similar fashion, $q(\ro_{t:T} | \rs_{t:T}, \theta, \pi)$ specifies beliefs about future rewards, given a policy. 

Given this factorization, it is straightforward to show that $-\mathcal{\tilde{F_{\pi}}}$ decomposes into an expected information gain term and an extrinsic term (see Appendix \ref{ap:feef})\footnote{The approximation in Eq. \ref{eq:decomposed} arises from the approximation $q(\rs_{0:T}, \theta | \ro_{0:T}, \pi) \approx p(\rs_{0:T}, \theta | \ro_{0:T}, \pi)$, which is justifiable given that $q(\cdot)$ represents a variational approximation of the true posterior \citep{friston2017active}.}:
\begin{equation}
\begin{aligned}
\label{eq:decomposed}
-\mathcal{\tilde{F_{\pi}}} \approx &-\underbrace{\E_{q(\ro_{0:T}|\pi)} \Big[ \KL \Big( q(\rs_{0:T}, \theta | \ro_{0:T}, \pi) \Vert q(\rs_{0:T}, \theta| \pi) \Big) \Big]}_\text{Expected information gain} \\
&+ \underbrace{\E_{q(\rs_{0:T},\theta | \pi)} \Big[ \KL \Big( q(\ro_{0:T} | \rs_{0:T}, \theta, \pi) \Vert p^{\bias}(\ro_{t:T}) \Big) \Big]}_\text{Extrinsic term}
\end{aligned}
\end{equation}

Maximizing Eq.\ref{eq:decomposed} has two functional consequences.
First, it maximises the expected information gain, which quantifies the amount of information an agent expects to gain from executing some policy. 
As agents maintain beliefs about the state of the environment and model parameters, this term promotes exploration in both state and parameter space. 

Second, it minimizes the extrinsic term - which is the KL-divergence between an agent's (policy-conditioned) beliefs about future observations and their preferred observations.
In the current context, it measures the KL-divergence between the rewards an agent expects from a policy and the rewards an agent desires.
In summary, selecting policies to minimise $\tilde{\mathcal{F}}$ invokes a natural balance between exploration and exploitation. 


\subsection{Relationship to probabilistic RL}
\label{relationship}

In recent years, there have been several attempts to formalize RL in terms of probabilistic inference \citep{levine2018reinforcement}, such as KL-control \citep{rawlik2013probabilistic}, control-as-inference \citep{kappen2012optimal}, and state-marginal matching \citep{lee2019efficient}. In many of these approaches, the RL objective is broadly conceptualized as minimising $\KL\Big(p(\ro_{0:T}|\pi) \Vert \ p^{\bias}(o_{0:T})\Big)$\footnote{We acknowledge that not all objectives follow this exact formulation.}.
In Appendix \ref{ap:bound-deriv}, we demonstrate that the free energy of the expected future $\mathcal{\tilde{F}}$ provides a tractable bound on this objective:
\begin{equation}
\begin{aligned}
   \mathcal{\tilde{F}} \geq \KL\Big(p(\ro_{t:T}|\pi) \Vert \ p^{\bias}(o_{t:T})\Big)
\end{aligned}
\end{equation}
These results suggest a deep homology between active inference and existing approaches to probabilistic RL. 


\section{Implementation}
\label{implementation}

In this section, we describe an efficient implementation of the proposed objective function in the context of model-based RL. 
To select actions, we optimise $q(\pi)$ at each time step, and execute the first action specified by the most likely policy. 
This requires (i) a method for evaluating beliefs about future variables $q(\rs_{t:T},\ro_{t:T}, \theta| \pi)$, (ii) an efficient method for evaluating $\mathcal{F}_{\pi}$, and (iii) a method for optimising $q(\pi)$ such that $q(\pi) = \sigma(-\mathcal{F}_{\pi})$


\paragraph{Evaluating beliefs about the future} We factorize and evaluate the beliefs about the future as:
\begin{equation}
\begin{aligned}
\label{eq:factor}
q(\rs_{t:T},\ro_{t:T}, \theta| \pi) &= q(\theta) \prod_{t=\tau}^T q(\ro_\tau| \rs_\tau, \theta, \pi) q(\rs_\tau| \rs_{\tau-1}, \theta, \pi) \\
q(\ro_\tau| \rs_\tau, \theta, \pi) &= \E_{q(\rs_\tau|\theta, \pi)} \big[p(\ro_\tau|\rs_\tau)\big] \\
q(\rs_\tau| \rs_{\tau-1}, \theta, \pi) &= \E_{q(\rs_{\tau-1}|\theta, \pi)} \big[p(\rs_\tau|\rs_{\tau-1}, \theta, \pi)\big]
\end{aligned}
\end{equation} 
where we have here factorized the generative model as $p(\ro_\tau, \rs_\tau, \theta| \pi) = p(\ro_\tau| \rs_\tau, \pi) p(\rs_\tau| \rs_{\tau-1}, \theta, \pi) p(\theta)$. We describe the implementation and learning of the likelihood $p(\ro_\tau| \rs_\tau, \pi)$, transition model $p(\rs_\tau| \rs_{\tau-1}, \theta, \pi)$ and parameter prior $p(\theta)$ in Appendix \ref{ap:model-details}. 


\paragraph{Evaluating $\mathcal{\tilde{F}_{\pi}}$}

Note that $-\mathcal{\tilde{F}}_{\pi} = \sum_{\tau=t}^{t + H} -\mathcal{\tilde{F}}_{\pi_{\tau}}$, where $H$ is the planning horizon. Given beliefs about future variables, the free energy of the expected future for a single time point can be efficiently computed as (see Appendix \ref{ap:exp-gain}):

\begin{equation}
\begin{aligned}
\label{eq:expected-free-energy-param}
   -\mathcal{\tilde{F}}_{\pi_{\tau}} &\approx E_{q(\rs_\tau,\theta | \pi)} \Big[ \KL \Big( q(\ro_\tau | \rs_\tau, \theta, \pi) \Vert p^{\bias}(\ro_\tau) \Big) \Big] \\ 
   &+ \underbrace{\mathbf{H}[q(\ro_{\tau}|\pi)] - \E_{q(\rs_{\tau}|\pi)}\Big[\mathbf{H}[q(\ro_{\tau}|\rs_{\tau}, \pi)]\Big]}_\text{State information gain}  \\ 
   &+ \underbrace{\mathbf{H}[q(\rs_{\tau}|\rs_{\tau-1}, \theta, \pi)] - \mathbb{E}_{q(\theta)}\Big[\mathbf{H}[q(\rs_{\tau}|\rs_{\tau-1}, \pi, \theta)]\Big]}_\text{Parameter information gain}
\end{aligned}	
\end{equation}

In the current paper, agents observe the true state of the environment $\rs_t$, such that the only partial observability is in rewards $\rr_t$. As as a result, the second term of equation \ref{eq:expected-free-energy-param} is redundant, as there is no uncertainty about states. The first (extrinsic) term can be calculated analytically (see Appendix \ref{ap:model-details}). We describe our approximation of the final term (parameter information gain) in Appendix \ref{ap:exp-gain}.



\paragraph{Optimising the policy distribution} We choose to parametrize $q(\pi)$ as a diagonal Gaussian. 
We use the CEM algorithm \citep{rubinstein1997optimization} to optimise the parameters of $q(\pi)$ such that $q(\pi) \propto -\mathcal{F}_{\pi}$.  
While this solution will fail to capture the exact shape of $-\mathcal{F}_{\pi}$, agents need only identify the peak of the landscape to enact the optimal policy. 


The full algorithm for inferring $q(\pi)$ is provided in Algorithm \ref{algo:exps}.

\begin{algorithm}[H]
\label{algo:exps}
\SetAlgoLined
   \DontPrintSemicolon
   \textbf{Input:} Planning horizon $H$ | Optimisation iterations $I$ | Number of candidate policies $J$ | Current state $\rs_t$ | Likelihood $p(\ro_\tau|\rs_\tau)$ |  Transition distribution $p(\rs_\tau|\rs_{\tau-1}, \theta, \pi)$ | Parameter distribution $P(\theta)$ | Global prior $p^{\bias}(\ro_\tau)$
   \BlankLine
   Initialize factorized belief over action sequences $q(\pi) \leftarrow \mathcal{N}(0,\mathbb{I})$.
   
   \For{$\mathrm{optimisation \ iteration} \ i = 1...I$}{
      Sample $J$ candidate policies from $q(\pi)$ \;
      \For{$\mathrm{candidate \ policy} \ j = 1...J$}{
      $\pi^{(j)} \sim q(\pi)$ \;
      $-\mathcal{\tilde{F}}_{\pi}^j = 0$ \;
      \For{$\tau = t...t+H$}{
         $q(\rs_\tau| \rs_{\tau-1}, \theta, \pi^{(j)}) = \E_{q(\rs_{\tau-1}|\theta, \pi^{(j)})} \big[p(\rs_\tau|\rs_{\tau-1}, \theta, \pi^{(j)})\big]$ \;
         $q(\ro_\tau| \rs_\tau, \theta, \pi^{(j)}) = \E_{q(\rs_\tau|\theta, \pi^{(j)})} \big[p(\ro_\tau|\rs_\tau)\big]$ \;
         $-\mathcal{\tilde{F}}_{\pi}^j \leftarrow -\mathcal{\tilde{F}}_{\pi}^j + E_{q(\rs_\tau,\theta | \pi^{(j)})} \big[ \KL \big( q(\ro_\tau | \rs_\tau, \theta, \pi^{(j)}) \Vert p^{\bias}(\ro_\tau) \big) \big]  
          +\mathbf{H}[q(\rs_{\tau}| \rs_{\tau-1}, \theta, \pi^{(j)})] - \mathbb{E}_{q(\theta)}\big[\mathbf{H}[q(\rs_{\tau}| \rs_{\tau-1}, \pi^{(j)}, \theta)]\big]$
      }
   }
   $q(\pi) \leftarrow \mathrm{refit}(-\mathcal{\tilde{F}}_{\pi}^j)$
}
\textbf{return} $q(\pi)$
\caption{Inference of $q(\pi)$}
\end{algorithm}


\section{Experiments}
\label{experiments}

To determine whether our algorithm successfully balances exploration and exploitation, we investigate its performance in domains with (i) well-shaped rewards, (ii) extremely sparse rewards and (iii) a complete absence of rewards. We use four tasks in total. For sparse rewards, we use the \textbf{Mountain Car} and \textbf{Cup Catch} environments, where agents only receive reward when the goal is achieved. For well-shaped rewards, we use the challenging \textbf{Half Cheetah} environment, using both the running and flipping tasks. For domains without reward, we use the \textbf{Ant Maze} environment, where there are no rewards and success is measured by the percent of the maze covered (see Appendix \ref{ap:env-details} for details on all environments).

\begin{figure}[h]
   \label{fig:a}
   \centering 
      \includegraphics[width=15cm]{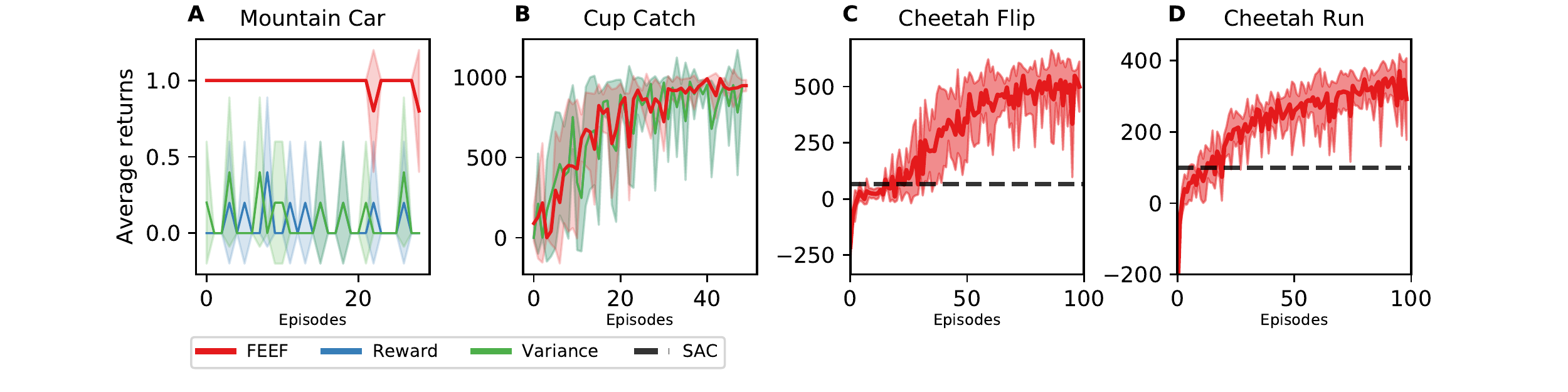}
      \caption{\textbf{(A) Mountain Car:} Average return after each episode on the sparse-reward Mountain Car task. Our algorithm achieves optimal performance in a single trial. \textbf{(B) Cup Catch:} Average return after each episode on the sparse-reward Cup Catch task. Here, results amongst algorithms are similar, with all agents reaching asymptotic performance in around 20 episodes. \textbf{(C \& D) Half Cheetah:} Average return after each episode on the well-shaped Half Cheetah environment, for the running and flipping tasks, respectively. We compare our results to the average performance of SAC after 100 episodes learning, demonstrating our algorithm can perform successfully in environments which do not require directed exploration. Each line is the mean of 5 seeds and filled regions show +/- standard deviation.}
   \end{figure}

For environments with sparse rewards, we compare our algorithm to two baselines, (i) a \textbf{reward} algorithm which only selects policies based on the extrinsic term (i.e. ignores the parameter information gain), and (ii) a \textbf{variance} algorithm that seeks out uncertain transitions by acting to maximise the output variance of the transition model (see Appendix \ref{ap:model-details}). Note that the variance agent is also augmented with the extrinsic term to enable comparison. For environments with well-shaped rewards, we compare our algorithm to the maximum reward obtained by a state-of-the-art model-free RL algorithm after 100 episodes, the soft-actor-critic (SAC) \cite{haarnoja2018soft}, which encourages exploration by seeking to maximise the entropy of the policy distribution. Finally, for environments without rewards, we compare our algorithm to a random baseline, which conducts actions at random.

The Mountain Car experiment is shown in Fig. 1A, where we plot the total reward obtained for each episode over 25 episodes, where each episode is at most 200 time steps. These results demonstrate that our algorithm rapidly explores and consistently reaches the goal, achieving optimal performance in a single trial. In contrast, the benchmark algorithms were, on average, unable to successfully explore and achieve good performance. We qualitatively confirm this result by plotting the state space coverage with and without exploration (Fig. 2B). Our algorithm performs comparably to benchmarks on the Cup Catch environment (Fig. 1B). We hypothesize that this is because, while the reward structure is technically sparse, it is simple enough to reach the goal with random actions, and thus the directed exploration afforded by our method provides little benefit. 

Figure 1 C\&D shows that our algorithm performs substantially better than a state of the art model-free algorithm after 100 episodes on the challenging Half Cheetah tasks. Our algorithm thus demonstrates robust performance in environments with well-shaped rewards and provides considerable improvements in sample-efficiency, relative to SAC.

Finally, we demonstrate that our algorithm can perform well in environments with no rewards, where the only goal is exploration. Figure 2B shows that our algorithms rate of exploration is substantially higher than that of a random baseline in the ant-maze environment, resulting in a more substantial portion of the maze being covered. This result demonstrates that the directed exploration afforded by minimising the free energy of the expected future proves beneficial in environments with no reward structure.

Taken together, these results show that our proposed algorithm - which naturally balances exploration and exploitation - can successfully master challenging domains with a variety of reward structures. 

\begin{figure}[h]
   \label{fig:b}
   \centering 
   \includegraphics[width=12cm, height=4cm]{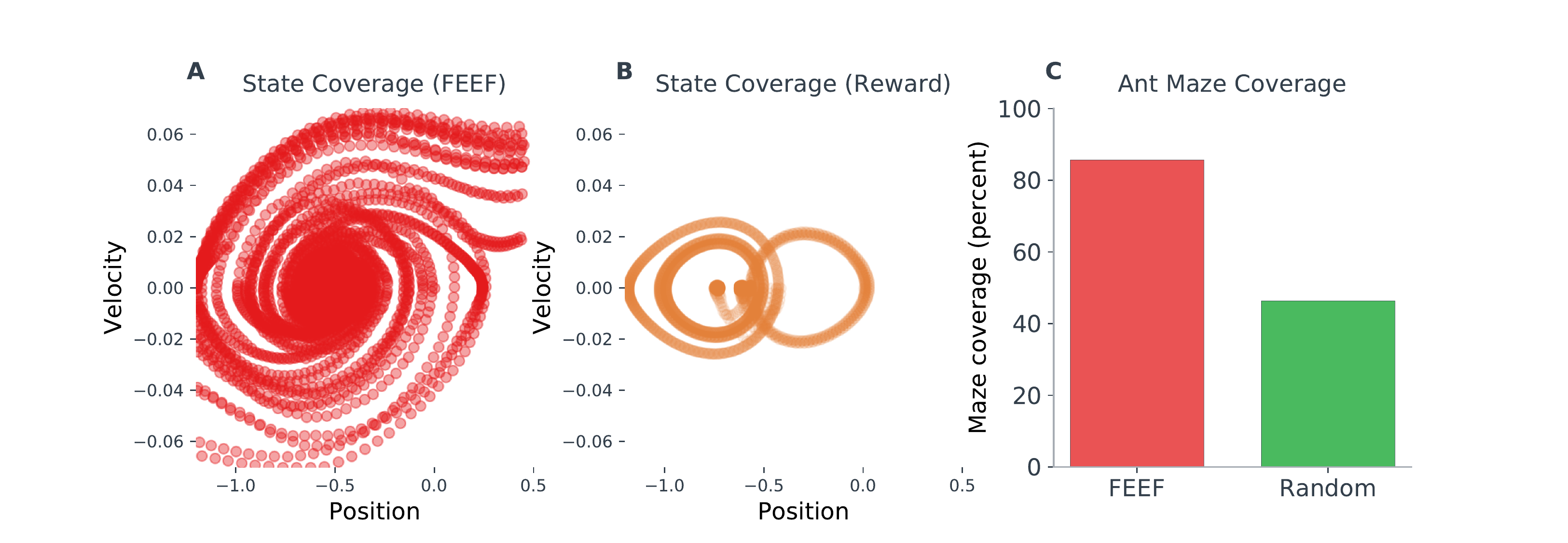}
\caption{\textbf{(A \& B) Mountain Car state space coverage}: We plot the points in state-space visited by two agents - one that minimizes the free energy of the expected future (FEEF) and one that maximises reward. The plots are from 20 episodes and show that the FEEF agent searches almost the entirety of state space, while the reward agent is confined to a region that be reached with random actions. \textbf{(C) Ant Maze Coverage}: We plot the percentage of the maze covered after 35 episodes, comparing the FEEF agent to an agent acting randomly. These results are the average of 4 seeds.}
\end{figure}


\section{Discussion} 
\label{discussion}

Despite originating from different intellectual traditions, active inference and RL both address fundamental questions about adaptive decision-making in unknown environments. Exploiting this conceptual overlap, we have applied an active inference perspective to the reward maximization objective of RL, recasting it as minimizing the divergence between desired and expected futures. We derived a novel objective that naturally balances exploration and exploitation and instantiated this objective within a model-based RL context. Our algorithm exhibits robust performance and flexibility in a variety of environments known to be challenging for RL. Moreover, we have shown that our algorithm applies to a diverse set of reward structures. Conversely, by implementing active inference using tools from RL, such as amortising inference with neural networks, deep ensembles and sophisticated algorithms for planning (CEM), we have demonstrated that active inference can scale to high dimensional tasks with continuous state and action spaces.

While our results have highlighted the existing overlap between active inference and RL, we end by reiterating two aspects of active inference that may be of utility for RL. First, representing preferences as a distribution over observations allows for greater flexibility in modelling and learning non-scalar and non-monotonic reward functions. This may prove beneficial when learning naturalistic tasks in complex nonstationary environments. Second, the fact that both intrinsic and extrinsic value are complementary components of a single objective - the free energy of the expected future - may suggest new paths to tackling the exploration-exploitation dilemma. Our method also admits promising directions for future work. These include investigating the effects of different distributions over reward, extending the approach to models which are hierarchical in time and space \citep{friston_deep_2018,PEZZULO2018294}, and investigating the deep connections to alternative formulations of probabilistic control.


\section*{Acknowledgements}

AT is funded by a PhD studentship from the Dr. Mortimer and Theresa Sackler Foundation and the School of Engineering and Informatics at the University of Sussex. 
BM is supported by an EPSRC funded PhDS Studentship.
CLB is supported by BBRSC grant number BB/P022197/1.
AT and AKS are grateful to the Dr. Mortimer and Theresa Sackler Foundation, which supports the Sackler Centre for Consciousness Science. 
AKS is additionally grateful to the Canadian Institute for Advanced Research  (Azrieli Programme on Brain, Mind, and Consciousness). 

\section*{Author Contributions}

A.T, B.M and C.L.B contributed to the conceptualization of this work. A.T and B.M contributed to the coding and generation of experimental results. A.T, B.M, C.L.B, A.K.S contributed to the writing of the manuscript.


\bibliographystyle{iclr2020_conference}
\bibliography{refs}
\newpage

\appendix

\section{Related work}
\label{ap:related_work}

\paragraph{Active inference} There is an extensive literature on active inference within discrete state-spaces, covering a wide variety of tasks, such as epistemic foraging in saccades \citep{parr2017active, friston_free_2019, schwartenbeck_computational_2019}, exploring mazes \citep{friston_active_2015,pezzulo2016active,friston2016active}, to playing Atari games \citep{cullen2018active}. Active inference also comes equipped with a well-developed neural process theory \citep{friston2017active,parr2019neuronal} which can account for a substantial range of neural dynamics. There have also been prior attempts to scale up active inference to continuous RL tasks \citep{tschantz_scaling_2019, millidge_deep_2019, ueltzhoffer_deep_2018}, which we build upon here.

\paragraph{Model based RL} Model based reinforcement learning has been in a recent renaissance, with implementations vastly exceeding the sample efficiency of model-free methods, while also approaching their asymptotic performance \citep{ha2018recurrent,nagabandi2018neural,chua2018deep,hafner2018learning}. There have been recent successes on challenging domains such as Atari \citep{kaiser_model-based_2019}, and high dimensional robot locomotion \citep{hafner2018learning,hafner2019dream} and manipulation \citep{nagabandi2019deep} tasks. Key advances include variational autoencoders \citep{kingma_auto-encoding_2013} to flexibly construct latent spaces in partially observed environments, Bayesian approaches such as Bayes by backprop \citep{houthooft2016curiosity}, deep ensembles \citep{shyam2018model,chua2018deep}, and other variational approaches \citep{okada_variational_2019, tschiatschek_variational_2018, yarin_gal_improving_2016}, which quantify uncertainty in the dynamics models, and enable the model to learn a latent space that is useful for action \citep{tschantz_learning_2019, watter_embed_2015}. Finally, progress has been aided by powerful planning algorithms capable of online planning in continuous state and action spaces \citep{williams2016aggressive,rubinstein1997optimization}.

\paragraph{Intrinsic Measures} Using intrinsic measures to encourage exploration has a long history in RL \citep{schmidhuber1991possibility,schmidhuber2007simple,storck1995reinforcement,oudeyer2009intrinsic,chentanez_intrinsically_2005}. Recent model-free and model based-intrinsic measures that have been proposed in the literature include policy-entropy \citep{rawlik2013probabilistic,rawlik2013stochastic,haarnoja2018soft},state entropy \citep{lee2019efficient}, information-gain \citep{houthooft_vime:_2016,okada_variational_2019,kim_emi:_2018,shyam_model-based_2019,teigen_active_2018}, prediction error \citep{pathak2017curiosity}, divergence of ensembles \citep{shyam_model-based_2019,chua_deep_2018}, uncertain state bonuses \citep{bellemare_unifying_2016,o2017uncertainty}, and empowerment \citep{de2018unified,leibfried2019unified,mohamed_variational_2015}. Information gain additionally has a substantial history outside the RL framework, going back to \citep{lindley_measure_1956,still_information-theoretic_2012,sun_planning_2011}.


\section{Derivation for the free energy of the expected future}
\label{ap:feef}

We begin with the full free energy of the expected future and decompose this into the free energy of the expected future given policies, and the negative policy entropy:
\begin{equation}
\begin{aligned}
   \mathcal{\tilde{F}} &= \E_{q(\ro, \rs, \theta, \pi)} [ \log q(\ro, \rs, \theta, \pi) - \log p^{\bias}(\ro, \rs, \theta) ] \\
   &= \E_{q(\pi)}[\mathcal{\tilde{F}}_{\pi}] - \mathbf{H}[q(\pi)]
\end{aligned}
\end{equation}

We now show the free energy of the expected future given policies can be decomposed into extrinsic and information gain terms:
\begin{equation}
\begin{aligned} 
      \mathcal{\tilde{F}}_{\pi} &= E_{q(\ro, \rs, \theta, \pi)} [ \log q(\ro, \rs, \theta, \pi) - \log p^{\bias}(\ro, \rs, \theta) ] \\
      &= \E_{q(\ro, \rs, \theta | \pi)} [ \log q(\rs, \theta| \pi) +  \log q(\ro | \rs, \theta, \pi) - \log p(\rs, \theta | \ro)  - \log p^{\bias}(\ro)  ] \\
      &\approx \E_{q(\ro, \rs, \theta | \pi)} [ \log q(\rs, \theta| \pi) +  \log q(\ro | \rs, \theta, \pi) - \log q(\rs, \theta | \ro, \theta)  - \log p^{\bias}(\ro)  ] \\
      &= \E_{q(\ro, \rs, \theta | \pi)} [ \log q(\rs, \theta| \pi) - \log q(\rs, \theta | \ro, \pi) ] + \E_{q(\ro, \rs, \theta | \pi)} [ \log q(\ro | \rs, \theta, \pi) - \log p^{\bias}(\ro) ] \\
      - \mathcal{\tilde{F}}_{\pi} &= \E_{q(\ro, \rs, \theta | \pi)} [  \log q(\rs, \theta | \ro, \pi)  - \log q(\rs, \theta |  \pi)] + \E_{q(\ro, \rs, \theta | \pi)} [ \log p^{\bias}(\ro) - \log q(\ro | \rs, \theta, \pi) ] \\
      &=  \underbrace{\E_{q(\ro|\pi)} \Big[ \KL \Big( q(\rs, \theta | \ro, \pi) \Vert q(\rs, \theta| \pi) \Big) \Big]}_{\text{Expected Information Gain}} - \underbrace{\E_{q(\rs, \theta | \pi)}\Big[ \KL \Big( q(\ro | \rs, \theta, \pi) \Vert p^{\bias}(\ro) \Big) \Big]}_{\text{Extrinsic Value}}
\end{aligned}
\end{equation}

Where we have assumed that $p(\rs, \theta|\ro) \approx q(\rs, \theta|\ro, \pi)$. We wish to minimize $\mathcal{\tilde{F}}_{\pi}$, and thus maximize $- \mathcal{\tilde{F}}_{\pi}$. This means we wish to maximize the information gain and minimize the KL-divergence between expected and preferred observations.

By noting that $q(\rs,\theta | \ro,\pi) \approx q(\rs | \ro,\pi)q(\theta | \rs)$, we can split the expected information gain term into state and parameter information gain terms:
\begin{equation}
    \begin{aligned}
        & \E_{q(\ro|\pi)} \Big[ \KL \Big( q(\rs, \theta | \ro, \pi) \Vert q(\rs, \theta| \pi) \Big) \Big] \\ 
        &= \E_{ q(\ro|\pi ) q(\rs, \theta | \ro, \pi)} \big[ \log q(\rs, \theta | \ro, \pi) - \log q(\rs, \theta| \pi) \big] \\
        &= \E_{q(\ro|\pi)q(\rs, \theta | \ro, \pi)} \big[ \log q(\rs | \ro, \pi) + \log q(\theta | \rs) - \log q(\rs | \theta, \pi) - \log q(\theta) \big] \\
        &= \E_{q(\ro|\pi)q(\rs, \theta | \ro, \pi)} \big[ \log q(\rs | \ro, \pi) - \log q(\rs | \theta, \pi) ]\big] +  \E_{q(\ro|\pi)q(\rs,\theta | \ro, \pi)} \big[ \log q(\theta | \rs)  - \log q(\theta) \big] \\
        &= \underbrace{\E_{q(\ro|\pi)q(\theta)} \Big[ \KL \big(q(\rs | \ro, \pi) \Vert q(\rs | \theta) \big) \Big]}_{\text{Expected State Information Gain}} + \underbrace{\E_{q(\rs | \theta)} \Big[\KL \big(q(\theta | \rs) \Vert q(\theta) \big) \Big]}_{\text{Expected Parameter Information Gain}}
    \end{aligned}
\end{equation}


\section{Derivation of the optimal policy}
\label{ap:opt-policy}

We derive the distribution for $q(\pi)$ which minimizes $\mathcal{\tilde{F}}$:

\begin{equation}
   \begin{aligned}
          \mathcal{\tilde{F}} &= \KL \Big( q(\ro, \rs, \theta , \pi ) \Vert p^{\bias}(\ro, \rs, \theta) \Big) \\
         &= \E_{q(\ro, \rs, \theta ,\pi)}[\log q(\ro, \rs, \theta |\pi) + \log q(\pi) - \log p^{\bias}(\ro, \rs, \theta,  \pi)] \\
         &=  \E_{q(\pi)} \Big[ \E_{q(\ro, \rs, \theta| \pi)}[ \log q(\pi) - [\log p^{\bias}(\ro, \rs, \theta) - \log q(\ro, \rs, \theta |\pi)]\Big] \\
         &= \E_{q(\pi)} \Big[\log q(\pi) - \E_{q(\ro, \rs, \theta| \pi)}[\log p^{\bias}(\ro, \rs, \theta) - \log q(\ro, \rs, \theta |\pi)]\Big] \\
         &= \E_{q(\pi)} \Big[\log q(\pi) - \big[-\E_{q(\ro, \rs, \theta| \pi)}[\log q(\ro, \rs, \theta |\pi) - \log p^{\bias}(\ro, \rs, \theta)]\big]\Big] \\
         &= \E_{q(\pi)} \Big[\log q(\pi) - \log e^{- \big[-\E_{q(\ro, \rs, \theta| \pi)}[\log q(\ro, \rs, \theta |\pi) - \log p^{\bias}(\ro, \rs, \theta)]\big]} \Big] \\
         &= \E_{q(\pi)} \Big[\log q(\pi) - \log e^{-\KL \big( q(\ro, \rs, \theta | \pi) \Vert p^{\bias}(\ro, \rs, \theta) \big)}\Big] \\
         &= \KL \Big( q(\pi)  \Vert  e^{-\KL \big( q(\ro, \rs, \theta | \pi) \Vert p^{\bias}(\ro, \rs, \theta) \big)} \Big) \\
         &= \KL \Big( q(\pi) \ \Vert  e^{-\mathcal{\tilde{F_{\pi}}}} \Big) \\
\end{aligned}
\end{equation}


\section{Derivation of RL bound}
\label{ap:bound-deriv}

Here we show that the free energy of the expected future is a bound on the divergence between expected and desired observations. The proof proceeds straightforwardly by importance sampling on the approximate posterior and then applying Jensen's inequality:
\begin{equation}
    \begin{aligned}
         \KL \big(q(\ro_{t:T} | \pi) \Vert \ p^{\bias}(o_{t:T})\big) &= \E_{q(\ro_{t:T}|\pi)}\big[ \log q(\ro_{t:T}|\pi) - \log p^{\bias}(\ro) \big] \\
         &= \E_{q(\ro_{t:T}|\pi)}\bigg[ \log \big( \int dx_{1:T} \int d\theta_{1:T} \frac{q(\ro_{t:T},\rs_{t:T},\theta_{t:T}|\pi)q(\rs_{t:T},\theta_{t:T} | \ro_{t:T})}{p^{\bias}(\ro_{t:T})q(\rs_{t:T},\theta_{t:T} | \ro_{t:T})} \big) \bigg] \\
         &\leq  \E_{q(\ro_{t:T},\rs_{t:T},\theta_{t:T}|\pi)} \Big[ \log \big( \frac{q(\ro_{t:T},\rs_{t:T},\theta_{t:T}|\pi)}{p^{\bias}(\ro_{t:T},\rs_{t:T},\theta_{t:T})} \big) \Big] \\
         &\leq  \KL \Big( q(\ro_{t:T}, \rs_{t:T}, \theta | \pi) \Vert p^{\bias}(\ro_{t:T}, \rs_{t:T}, \theta) \Big) = \mathcal{\tilde{F}}
    \end{aligned}
\end{equation}


\section{Model details}
\label{ap:model-details}

In the current work, we implemented our probabilistic model using an ensemble-based approach \citep{chua2018deep,fort_deep_2019,chitta_deep_2018}. Here, an ensemble of point-estimate parameters $\theta = \{\theta_0 , ... , \theta_B\}$ trained on different batches of the dataset $\mathcal{D}$ are maintained and treated as samples as from the posterior distribution $p(\theta|\mathcal{D})$. Besides consistency with the active inference framework, probabilistic models enable the active resolution of model uncertainty, capture both epistemic and aleatoric uncertainty, and help avoid over-fitting in low data regimes \citep{fort_deep_2019, chitta_deep_2018, chatzilygeroudis_survey_2018, chua_deep_2018}. 

This design choice means that we use a trajectory sampling method when evaluating beliefs about future variables \citep{chua2018deep}, as each pass through the transition model $p(\rs_t| \rs_{t-1}, \theta, \pi)$ evokes $B$ samples from $\rs_t$. 

\paragraph{Transition model} We implement the transition model as $p(\rs_t| \rs_{t-1}, \theta, \pi)$ as $\mathcal{N}(\rs_t; f_{\theta}(\rs_{t-1}),f_{\theta}(\rs_{t-1}))$, where $f_{\theta}(\cdot)$ are a set of function approximators $f_{\theta}(\cdot) = \{f_{\theta_0}(\cdot) , ... , f_{\theta_B}(\cdot)\}$. In the current paper, $f_{\theta_i}(\rs_{t-1})$ is a two-layer feed-forward network with 400 hidden units and swish activation function. Following previous work, we predict state deltas rather than the next states \citep{shyam2018model}.

\paragraph{Reward model} We implement the reward model as $p(\ro_\tau | \rs_\tau, \theta, \pi) = \mathcal{N}(\ro_\tau; f_{\lambda}(\rs_\tau), \mathbf{1})$, where $f_{\lambda}(\rs_\tau)$ is some arbitrary function approximator\footnote{Formally, this is an observation model, but we retain RL terminology for clarity.}. In the current paper, $f_{\lambda}(\rs_\tau)$ is a two layer feed forward network with 400 hidden units and ReLU activation function. Learning a reward model offers several plausible benefits outside of the active inference framework, as it abolishes the requirement that rewards can be directly calculated from observations or states \citep{chua2018deep}.

\paragraph{Global prior} We implement the global prior $p^{\bias}(\ro)$ as a Gaussian with unit variance centred around the maximum reward for the respective environment. We leave it to future work to explore the effects of more intricate priors.


\section{Implementation details}
\label{ap:implem-details}

For all tasks, we initialize a dataset $\mathcal{D}$ with a single episode of data collected from a random agent. For each episode, we train the ensemble transition model and reward model for 100 epochs, using the negative-log likelihood loss. We found cold-starting training at each episode to lead to more consistent behaviour. We then let the agent act in the environment based on Algorithm \ref{algo:exps}, and append the collected data to the dataset $\mathcal{D}$.

We list the full set of hyperparameters below:
\begin{center}
    
   \begin{tabular}{ |p{5cm}||p{3cm} | }
    \hline
    \multicolumn{2}{|c|}{Hyperparameters} \\
    \hline
    Hidden layer size   & 400   \\
    Learning rate&   0.001  \\
    Training-epochs & 100\\
    Planning-horizon & 30 \\
    N-candidates (CEM) &  700  \\
    Top-candidates (CEM) & 70  \\
    Optimisation-iterations (CEM) & 7 \\
    \hline
   \end{tabular}
   \end{center}


\section{Expected information gain}
\label{ap:exp-gain}

In Eq. \ref{eq:decomposed}, expected parameter information gain was presented in the form $\E_{q(\rs | \theta)}\KL \big(q(\theta | \rs) \Vert q(\theta) \big)$. While this provides a nice intuition about the effect of the information gain term on behaviour, it cannot be computed directly, due to the intractability of identifying true posteriors over parameters. We here show that, through a simple application of Bayes' rule, it is straightforward to derive an equivalent expression for the expected information gain as the divergence between the state likelihood and marginal, given the parameters, which decomposes into an entropy of an average minus an average of entropies:
\begin{equation}
    \begin{aligned}
        & \E_{q(\rs | \theta)}\KL \big(q(\theta | \rs) \Vert q(\theta) \big) \\
        &= \E_{q(\rs | \theta)q(\theta | \rs)}\big [ \log q(\theta | \rs) - \log q(\theta) \big] \\
        &= \E_{q(\rs, \theta)}\big [ \log q(\rs | \theta) + \log q(\theta) - \log q(\rs) - \log q(\theta) \big] \\
        &= \E_{q(\rs, \theta)}\big [ \log q(\rs | \theta)  - \log q(\rs) \big] \\
        &= \E_{q(\theta)q(\rs | \theta)}\big [\log q(\rs | \theta) \big]  -\E_{q(\theta)q(\rs|\theta)} \big[ \log \E_{q(\theta)} q(\rs | \theta) \big] \\
        &= -\E_{q(\theta)}\mathbf{H}\big[ q(\rs | \theta) \big] + \mathbf{H} \big[ \E_{q(\theta)}q(\rs | \theta) \big]
    \end{aligned}
\end{equation}

The first term is the (negative) average of the entropies. The average over the parameters $\theta$ is achieved simply by averaging over the dynamics models in the ensemble. The entropy of the likelihoods $\mathbf{H}[p(\rs|\theta)]$ can be computed analytically since each network in the ensemble outputs a Gaussian distribution for which the entropy is a known analytical result. The second term is the entropy of the average $\mathbf{H}[\mathbb{E}_{p(\theta)}p(\rs|\theta)]$. Unfortunately, this term does not have an analytical solution. However, it can be approximated numerically using a variety of techniques for entropy estimation. In our paper, we use the nearest neighbour entropy approximation \citep{mirchev_approximate_2018}.


\section{Environment details}
\label{ap:env-details}
The Mountain Car environment ($\mathcal{S} \subseteq \mathbb{R}^2 \mathcal{A} \subseteq \mathbb{R}^1$) requires an agent to drive up the side of a hill, where the car is underactuated requiring it first to gain momentum by driving up the opposing hill. A reward of one is generated when the agent reaches the goal, and zero otherwise. The Cup Catch environment ($\mathcal{S} \subseteq \mathbb{R}^8 \mathcal{A} \subseteq \mathbb{R}^2$) requires the agent to actuate a cup and catch a ball attached to its bottom. A reward of one is generated when the agent reaches the goal, and zero otherwise. The Half Cheetah environment ($\mathcal{S} \subseteq \mathbb{R}^17 \mathcal{A} \subseteq \mathbb{R}^6$) describes a running planar biped. For the running task, a reward of $v - 0.1 ||a||^2$ is received, where $v$ is the agent's velocity, and for the flipping task, a reward of $\epsilon - 0.1 ||a||^2$ is received, where $\epsilon$ is the angular velocity. The Ant Maze environment ($\mathcal{S} \subseteq \mathbb{R}^29 \mathcal{A} \subseteq \mathbb{R}^8$) involves a quadruped agent exploring a rectangular maze. 

\end{document}

%% file: math_commands.tex
\usepackage{amsmath,amsfonts,bm}

\def\eqref#1{equation~\ref{#1}}

\def\1{\bm{1}}

\def\ro{{\textnormal{o}}}

\def\rr{{\textnormal{r}}}
\def\rs{{\textnormal{s}}}

\def\rx{{\textnormal{x}}}

\DeclareMathAlphabet{\mathsfit}{\encodingdefault}{\sfdefault}{m}{sl}
\SetMathAlphabet{\mathsfit}{bold}{\encodingdefault}{\sfdefault}{bx}{n}

\newcommand{\E}{\mathbb{E}}

\newcommand{\KL}{D_{\mathrm{KL}}}

